\documentclass[journal,letterpaper]{IEEEtran}

\usepackage[noadjust]{cite}

\usepackage[utf8]{inputenc}
\usepackage[T1]{fontenc}   
\usepackage{hyperref}       
\usepackage{url}           
\usepackage{booktabs}       
\usepackage{amsfonts}       
\usepackage{nicefrac}       
\usepackage{microtype}

\usepackage{amssymb,bm}
\usepackage{amsthm,algorithm,algorithmic,bbm,mathtools,amsmath}

\newcommand{\bs}[1]{\boldsymbol{#1}}
\newcommand{\vt}{{\bs{\theta}}}

\newcommand{\vx}{\bs{x}}

\newcommand{\sP}{{\cal P}}

\newcommand{\Tau}{\mathcal{T}}
\newcommand{\Wau}{\mathcal{W}}
\newcommand{\Hau}{\mathcal{H}}
\newcommand{\Nau}{\mathcal{N}}

\newtheorem{theorem}{Theorem}
\newtheorem{lemma}[]{Lemma}
\newtheorem{corollary}[]{Corollary}
\newtheorem{remark}[]{Remark}
\newtheorem{definition}[]{Definition}
\newtheorem{proposition}[]{Proposition}

\title{Optimal and Efficient Algorithms for General Mixable Losses against Switching Oracles}

\author{
	\IEEEauthorblockN{Kaan Gokcesu}, \IEEEauthorblockN{Hakan Gokcesu}
  }

\begin{document}

\maketitle

\begin{abstract}
  We investigate the problem of online learning, which has gained significant attention in recent years due to its applicability in a wide range of fields from machine learning to game theory. Specifically, we study the online optimization of mixable loss functions in a dynamic environment. We introduce online mixture schemes that asymptotically achieves the performance of the best dynamic estimation sequence of the switching oracle with optimal regret redundancies. The best dynamic estimation sequence that we compete against is selected in hindsight with full observation of the loss functions and is allowed to select different optimal estimations in different time intervals (segments). We propose two mixtures in our work. Firstly, we propose a tractable polynomial time complexity algorithm that can achieve the optimal redundancy of the intractable brute force approach. Secondly, we propose an efficient logarithmic time complexity algorithm that can achieve the optimal redundancy up to a constant multiplicity gap. Our results are guaranteed to hold in a strong deterministic sense in an individual sequence manner.
\end{abstract}

\section{Introduction}
We study the online learning problem \cite{cesa_book, poor_book}, which has applications in a wide variety of fields including decision theory \cite{tnnls4}, game theory \cite{tnnls1,chang}, control theory \cite{tnnls3}, signal processing \cite{ozkan}, multi-agent systems \cite{vanli}, optimization \cite{zinkevich,hazan}, density estimation and source coding \cite{gDensity,willems,coding1,gAnomaly,coding2}, anomaly and outlier detection \cite{gokcesu_anomaly2,gIncremental}, adversarial bandits \cite{cesa-bianchi,cesa2007,gBandit} and prediction \cite{singer,singer2,gokcesu_prediction}.

\subsection{Online Learning}
In the setting of the online learning problem, we make a estimation (e.g., prediction, action, belief) $\vt_t$ at each time $t$ and we receive the observations $\bs{x}_t$ \cite{cesa_book}. We incur a loss based on our estimation $\vt_t$ and the observation $\bs{x}_t$, i.e., $l(\vt_t,\bs{x}_t)$. Note that the loss functions need not be the same and at each time $t$ there may be different loss functions present. Hence, the loss at time $t$ can be given by $l_t(\vt_t,\bs{x}_t)$. In general, the problem is to produce estimations $\vt_t$, using the observed samples and losses up to time $t-1$ (i.e., $\{\bs{x}_\tau\}_{\tau=1}^{t-1}$ and $\{l_\tau(\vt_\tau,\bs{x}_\tau)\}_{\tau=1}^{t-1}$), and minimize the total loss incurred up to time $T$ (i.e., $L_T\triangleq\sum_{t=1}^T l_t(\vt_t,\bs{x}_t)$).
Hence, in the online learning setting, the goal is to produce an estimation $\vt_t$ without seeing (or knowing) observation $\bs{x}_t$ or the loss function $l_t(\cdot,\cdot)$ beforehand. This phenomena can be better understood by the online portfolio management problem \cite{li2014}. In the online portfolio management, the equity at hand is distributed among several stocks at time $t$. However, we cannot see the reward (or the loss) of this action until we see the price changes in the stock market. Thus, this phenomenon creates a time dependent loss problem. 

\subsection{Types of Losses}\label{sec:loss}
\subsubsection{Convex Loss}\label{sec:convex}
Because of its general applicability and straightforward solutions, the most popular loss function of interest in online learning literature is the convex loss \cite{boyd2004}. The losses are convex in its argument $\vt_t$ if $\bs{v}^T\left(\nabla^2_{\vt}l_t(\vt_t,\bs{x}_t)\right)\bs{v}\geq 0$ for every $\bs{v}$ (i.e., the Hessian should be positive semi-definite). While this loss is generally applicable because of its milder conditions, it has a somewhat slower learning rate.
\subsubsection{Strongly-Convex Loss}\label{sec:strong}
For faster convergence in the convex optimization, the stricter condition of strong convexity is studied \cite{strong_convex}. The loss is $\beta$-strongly-convex if it is convex in its argument $\vt_t$ such that $\bs{v}^T\left(\nabla^2_\vt l_t(\vt_t,\bs{x}_t)-\beta I\right)\bs{v}\geq 0$ for every $\bs{v}$ (i.e., the difference of Hessian and $\beta$ times the identity matrix should be positive semi-definite). While this loss has a much stronger learning rate, its conditions are too strict which limits its applicability, since in most well-known problems in the optimization literature, the strong convexity on the losses are not present \cite{cesa_book,poor_book}.
\subsubsection{Exp-Concave Loss}\label{sec:expconcave}
The conditions of the strongly-convex loss functions are loosened by the definition of exp-concave loss functions \cite{koren2013}.
The loss is $\lambda$-exp-concave if
\begin{align*}
\exp\left({-\lambda l_t\left(\sum_{i}{P}_i\bs{\theta}_i,\bs{x}\right)}\right)\geq\sum_{i}{P}_i\exp\left({-\lambda l_t\left(\bs{\theta}_i,\bs{x}\right)}\right)
\end{align*}
where $\{P_i\}_{i=1}^n$ is a probability distribution. This has similar learning rates as and milder conditions than strong convexity.
\subsubsection{Mixable Losses}\label{sec:mixable}
Because of it applicability, we focus on the mixable losses \cite{vovk}, where $l(\bs{\theta},\bs{x})$ is $\alpha$-mixable if the following definition holds.
\begin{definition}\label{def:mixable}
	$l(\bs{\theta},\bs{x})$ is $\alpha$-mixable in $\bs{\theta}$ if there exists a surrogate function $F(\cdot,\cdot)$ such that
	\begin{align*}
	\hat{\bs{\theta}}=F(\{\bs{\theta}_i,P_i\}_{i=1}^n),
	\end{align*}
	where $\{P_i\}_{i=1}^n$ is some probability distribution; and we have
	\begin{align*}
	e^{-\alpha l(\hat{\bs{\theta}},\bs{x})}\geq\sum_{i}{P}_ie^{-\alpha l(\bs{\theta}_i,\bs{x})}.
	\end{align*} 
\end{definition}
The surrogate function or the 'mixture rule' $F(\cdot)$ in Definition \ref{def:mixable} is specific for different mixable losses. As an example, the square loss function, where $\vt\in[-1,1]$, and $l(\vt,\vx)=(\vt-\vx)^2$ is mixable with $\alpha=\frac{1}{2}$ \cite{vovk, haussler1998}, where $\hat{\vt}$ is given by
\begin{align}
&\hat{\vt}=\frac{1}{2}\sum_{q=-1}^{1}q\log\left(\sum_{i=1}^{n}P_ie^{-\frac{1}{2}(\vt_i-q)^2}\right)
\end{align}
As can be seen, the exp-concave loss functions are a subclass of the mixable losses, by setting the surrogate function as the mean of $\bs{\theta}_i$ by $P_i$.

\subsection{Notion of Regret and Oracle}
The best scenario in our learning is to know the losses in advance and make the decisions accordingly. Since this is infeasible, our challenge is to achieve the performance of an algorithm that knows the loss functions beforehand, i.e., the oracle.
Moreover, since the losses can be arbitrarily high and achieving small losses may be difficult, achieving a loss as small as a viable competition is more feasible \cite{cesa_book}.
To this end, we work in a competitive framework and use the notion of regret. In its most general form, the regret against an arbitrary competition is defined as the difference between the losses we and the competition scheme incur. Against a competition $\{\vt^*_t\}_{t=1}^T$, the cumulative regret up to time $T$ is
\begin{align}
R_T\triangleq\sum_{t=1}^T r_t=\sum_{t=1}^T l_t(\vt_t,\bs{x}_t)-\sum_{t=1}^T l_t(\vt^*_t,\bs{x}_t).\label{eq:Rt}
\end{align}
\subsubsection{Competing Against Best Fixed Oracle}\label{sec:fixed}
Since competing against the scheme that chooses the optimal parameter at each time $t$ as in \eqref{eq:Rt} is infeasible, the traditional goal was to compete against the best fixed selection in hindsight as
\begin{align}
R_{T,F}\triangleq\sum_{t=1}^T l_t(\vt_t,\bs{x}_t)-\sum_{t=1}^T l_t(\vt^*,\bs{x}_t).
\end{align}
\subsubsection{Competing Against Best Switching Oracle}
A more challenging problem is to compete against a switching sequence of selections $\{\vt^*_s\}_{s=1}^S$ instead of a fixed selection ${\bs{\theta}^*}$, where $S$ is the number of times the competition selection changes.  
Each $\vt^*_s$ is individually optimal in their $S$ respective distinct and mutually exclusive time segments $\{T_s+1,T_s+2,\ldots,T_{s+1}\}$ of length $t_s$, where $t_s\triangleq T_{s+1}-T_s$, $T_1=0$ and $T_{S+1}=T$ (note that $\sum_{s=1}^{S}t_s=T$), i.e., we want to minimize the following:
\begin{align}
R_{T,S}\triangleq\sum_{s=1}^{S}\sum_{t=T_{s}+1}^{T_{s+1}}l(\vt_t,\vx_t)-l(\vt^*_s,\vx_t).\label{probS}
\end{align}
\subsection{Problem Definition and Literature Review}
Competing against the best fixed selection is well studied in the literature, which addresses the problem in Section \ref{sec:fixed}. In general, we have an algorithm that produces predictions $\vt_t^{B}\in \sP$, which has the following performance bound.
\begin{definition}\label{def:RB}
	The base algorithm and its estimations $\vt_t^B$ has
	\begin{align*}
	\sum_{t=1}^{T}l(\vt_t^B,\vx_t)-l(\vt^*,\vx_t)\leq&R_B(T),
	\end{align*}
	regret in a time horizon $T$, against $\vt^*$ with observations $\vx_t$.
\end{definition}
For the general convex losses in Section \ref{sec:convex}, the first order algorithms \cite{zinkevich} achieve $R_B(T)=O(\sqrt{T})$ against the best fixed estimation in hindsight (which is minimax optimal \cite{abernethy}) with fixed computational complexity per iteration. In \cite{hazan}, it is shown that we can achieve a lower $R_B(T)=O(\log(T))$ regret with the assumption of $\beta$-strong convexity in Section \ref{sec:strong} (by tuning the learning rates). In \cite{hazan}, the authors propose an algorithm that achieves $R_B(T)=O(\log(T))$ against the best fixed estimation with fixed (albeit higher) computational complexity per iteration for $\lambda$-exp-concave loss functions in Section \ref{sec:expconcave}. 
For the mixable losses in Section \ref{sec:mixable}, the weighting algorithms to aggregate all possible samples on the set such as \cite{vovk} can be used to achieve $R_B(T)=O(\log(T))$ regret against the best fixed estimation, which is computationally too demanding. An efficient solver for general mixable losses is not straightforward, because of their varying estimators, henceforth, they are generally tailored to the loss functions at hand \cite{cesa_book,haussler1998,vovk2001}.

While competing against the best switching selection can be directly achieved for the general convex losses in Section \ref{sec:convex} via optimizing the learning rate \cite{zinkevich}, it is not straightforward for the other losses in Section \ref{sec:loss}. If we were to know the time instances that the oracle changes its parameter selections, we can directly apply the algorithms that can solve the static problem (such as \cite{zinkevich,hazan,vovk}) in the given time segments separately to achieve minimax regret $SR_B(T/S)$. 
To solve the dynamic problem without a priori knowledge, certain techniques are utilized in literature that incorporate the idea of creating hyper-experts (which are created from the algorithms that can compete against the best fixed estimator) with appropriate selection of time intervals and mixing their estimations. 
The works in \cite{chernov2009,freund1997using} create a pool of experts each of which abstains from prediction at first and produces predictions starting from some trial $t$. In \cite{hazan2009}, they start with the base algorithm and restart a copy of it each trial. It is shown that both of these approaches reduce to the same algorithm with variable parameters \cite{adamskiy}. These approaches can be generalized to use different time selection functions as well \cite{blum2007}. \cite{gyorgy2012} and \cite{zhang2018} have developed ways to construct intervals which can trade effectiveness for efficiency explicitly. Especially, the exponential interval (which stems from the doubling trick) \cite{daniely,zhang2019,zhang2020} has become popular.
Techniques like exponential weights \cite{gokcesu2020recursive,auerExp,comp2} and multiplicative weights \cite{daniely} are used to mix these experts \cite{littlestone1994}. In the aggregation techniques, the goal is to minimize the redundancy without sacrificing modeling power and computational complexity \cite{gokcesu2020generalized}.
\subsection{Contributions and Organization}
\subsubsection{Section \ref{sec:pre}}
We provide some useful preliminaries; which include the algorithmic framework (that generalizes and complements the literature), some important definitions.
\subsubsection{Section \ref{sec:exp}}
We provide a brute force approach that can solve the switching problem. While the mixture is intractable, it provides a benchmark for regret comparison that coincides with the optimal achievable results in literature such as \cite{coding1}. 
\subsubsection{Section \ref{sec:quad}}
We provide a tractable approach (polynomial complexity) that can solve the switching problem exactly, albeit with some mixture redundancy. While the traditional mixtures like \cite{willems} can achieve near-optimal mixture redundancies, we construct a weighting scheme with optimal redundancy. 
\subsubsection{Section \ref{sec:log}}
We provide a practical approach (logarithmic complexity) that can solve the switching problem approximately (which is, in general, unavoidable for unknown $T$). While the mixtures of \cite{hazan2009} have redundancies with divergent optimality gap, our scheme has a constant optimality gap.   
\subsubsection{Section \ref{sec:con}}
We finish with some concluding remarks.
\section{Preliminaries}\label{sec:pre}
\subsection{Algorithm Framework}\label{sec:mix}
Suppose we have a set of $\Nau_t$ parallel running algorithms at time $t$. At each time $t$, each algorithm $i\in\Nau_t$ provides us with its parameter estimate $\bs{\theta}_{i,t}$. To create our estimate $\bs{\theta}_t$, we combine the provided estimates $\bs{\theta}_{i,t}$ with probabilities $P_{i,t}$ (where $\sum_{i\in\Nau_t}^{} P_{i,t}=1$) using the surrogate function $F(\cdot)$ as
\begin{align}
	\hat{\bs{\theta}}_t=F(\{\bs{\theta}_{i,t},P_{i,t}\}_{i\in\Nau_t}^{}).\label{thetahat}
\end{align}
Each $P_{i,t}$ is created by normalizing their weights $\tilde{P}_{i,t}$ as
\begin{align}
P_{i,t} &= \frac{\widetilde{P}_{i,t}}{\sum_{i\in\Nau_t}^{} \widetilde{P}_{i,t}}, && \text{for $i\in\Nau_t$}.\label{Pit}
\end{align}
We design the performance weights $\tilde{P}_{i,t}$ to be dependent on the performance of the algorithms. However, to compete against a dynamic strategy, we also share the weights of the experts with each other accordingly. Thus, we choose the performance weights such that they are recursively given by
\begin{align}
\widetilde{P}_{j,t} &= \sum_{i\in\Nau_{t-1}}^{} \widetilde{P}_{i,t-1} e^{-\alpha l_{t-1}(\bs{\theta}_{i,t-1})} \tau_t(i,j)\label{PitRec}
\end{align}
where $\tau_t(i,j)$ is the transition from the $i^{th}$ to $j^{th}$ algorithm.

\begin{definition}\label{def:Tau}
	We collectively define the transition weights $\tau_t(i,j)$ for all $(i,j,t)$ as a weighting scheme $\Tau$ such that
	\begin{align}
		\Tau:\enspace \tau_{t}(i,j), &&\forall i,j,t,
	\end{align}
	where $\tau_t(i,j)$ is nonnegative for all $(i,j,t)$ and
	\begin{align}
	\sum_{j}^{}\tau_t(i,j)\leq1, &&\forall j, t,
	\end{align}
	i.e., $\tau_t(i,\cdot)$ is upper-bounded by a probability distribution. 
\end{definition}

\begin{definition}\label{def:Hau}
	The experts $i$ are created from the base algorithm according to an hyper-expert scheme $\Hau$ such that
	\begin{align}
	\Hau: H(i)=\{\bs{\lambda}_i\}, &&\forall i,t,
	\end{align} 
	where $\bs{\lambda}_i$ collectively defines the necessary information about how to utilize the base algorithm for the $i^{th}$ expert (e.g., start and finish times of the individual runs of the base algorithm). 
\end{definition}

\begin{theorem}\label{thm:mixTrans}
	When the mixture uses the weighting scheme $\Tau$, which defines the transition weights $\tau_t(\cdot,\cdot), \enspace\forall t$; we have the following upper bound on our losses in terms of the losses of an arbitrary sequence of experts $\{I_1,I_2,\ldots,I_T\}$ from the hyper-expert scheme $\Hau$
	\begin{align}
	\sum_{t=1}^T l_t(\hat{\bs{\theta}}_t) \leq\sum_{t=1}^T l_t(\bs{\theta}_{I_t,t})+\frac{1}{\alpha}\Wau_\Tau\left(\{I_t\}_{t=1}^T\right).\nonumber
	\end{align}
	where $\Wau_\Tau\left(\{I_t\}_{t=1}^T\right)\triangleq-\log\left(\prod_{t=1}^T \tau_t(I_{t-1},I_t)\right)$, $I_t\in\Nau_t$, $\tau_1(I_0,I_1)\leq P_{I_1,1}$ and each $l_t(\cdot)$ is $\alpha$-mixable.
	\begin{proof}
		The proof is in Appendix \ref{app:lem:mixTrans}.
	\end{proof}
\end{theorem}

\subsection{Important Definitions}
\begin{definition}\label{def:S}
	In an $S$ segment competition, where we compete against parameters $\{\vt_s^*\}_{s=1}^S$ with time lengths $\{t_s\}_{s=1}^S$, let 
	\begin{align}
	\mathcal{I}(\{t_s\}_{s=1}^S)=\{I_t\}_{t=1}^T
	\end{align}
	be a sequence of experts from $\Hau$ that is able to 'compete'. This sequence of experts $\{I_t\}_{t=1}^T$ collectively imply a sequence of runs of the base algorithm (the structure of which depends on $\Hau$), where during each individual run, the competition $\vt_{s}^*$ stays the same. 
	Let $S_E$ be the number of segments we have in $\{I_t\}_{t=1}^T$ such that
	\begin{align}
	\mathcal{S}(\{t_s\}_{s=1}^S)\triangleq S_E= 1+\sum_{t=2}^{T}\mathbbm{1}_{I_t\neq I_{t-1}},\nonumber
	\end{align}
	where $\mathbbm{1}_x$ is the identity operator. We point out that $S_E\leq o(T)$ for viable learning, where $o(\cdot)$ is the Little-O notation. 
\end{definition}

\begin{definition}\label{def:Es}
	For a sequence of competition predictions $\{\vt_s^*\}_{s=1}^S$ with $S$ segments and the hyper-expert construction scheme $\Hau$, we define the following
	\begin{align}
	E_{S,T}(\mathcal{H})\triangleq\max_{\{t_s\}_{s=1}^S}\sum_{s=1}^{S}\sum_{t=T_{s-1}+1}^{T_s} l({\bs{\theta}}_{I_t,t},\vx_t)-l({\bs{\theta}}_s^*,\vx_t)\nonumber
	\end{align}
	which we call as the 'expert regret' of $\Hau$ for $S$ and $T$ (where $I_t$ comes from $\mathcal{H}$ as in Definition \ref{def:S}).
\end{definition}

\begin{definition}\label{def:Ws}
	In an $S$ segment competition, where we compete against parameters $\{\vt_s^*\}_{s=1}^S$ with time lengths $\{t_s\}_{s=1}^S$, for the expert scheme $\Hau$ and the weighting scheme $\Tau$, we define
	\begin{align}
	W_{S,T}(\Tau)\triangleq\max_{\{t_s\}_{s=1}^S}\Wau_\Tau\left(\mathcal{I}\left(\{t_s\}_{s=1}^S\right)\right),\nonumber
	\end{align}
	which we call as the 'mixture regret' of $\Tau$ for $S$ and $T$.
\end{definition}

\begin{corollary}
	We split the regret resulting from Theorem \ref{thm:mixTrans} as
	\begin{align*}
		R_{S,T}\left(\{\hat{\bs{\theta}}_t\}_{t=1}^T\right)
		\triangleq&\sum_{s=1}^{S}\sum_{t=T_{s-1}+1}^{T_s} l(\hat{\bs{\theta}}_t,\vx_t)-l({\bs{\theta}}_s^*,\vx_t)\\
		\leq& E_{S,T}(\mathcal{H})+\frac{1}{\alpha}W_{S,T}(\Tau),
	\end{align*}
	where $E_{S,T}(\mathcal{H})$ is the regret resulting from the hyper-expert construction $\Hau$; and $W_{S,T}(\Tau)$ is the regret redundancy resulting from the mixture weighting scheme $\Tau$ in a time horizon $T$ when competing against $S$ time segments.
\end{corollary}

\begin{definition}\label{def:approx}
	We use the following expression
	\begin{align*}
		A(S,T)\lessapprox B(S,T),
	\end{align*}
	to denote an asymptotic approximate bound, if, as $T\rightarrow\infty$,
	\begin{align*}
		A(S,T)\leq (1+\epsilon)B(S,T),
	\end{align*}
	for every finite $\epsilon >0$, i.e.,
	\begin{align*}
	A(S,T)\leq B(S,T)+o(B(S,T)).
	\end{align*}
\end{definition}

\begin{lemma}\label{lem:binom}
	We have the following inequalities
	\begin{align}
	\left(\frac{T}{S}\right)^S\leq\binom{T}{S}\leq\left(\frac{eT}{S}\right)^S,\label{binom}
	\end{align}
	which bounds a binomial coefficient.
	\begin{proof}
		The proof comes from \cite{farhi2007}.
	\end{proof}
\end{lemma}

\section{A~Brute~Force~Approach: An~Exponential~Time~Complexity~Mixture~Scheme}\label{sec:exp}
The easiest way to minimize the expression in \eqref{probS}, would be to know the time instances the optimal parameter changes (which are $\{T_s\}_{s=1}^S$) and restart the base algorithms after these times, which would have resulted in the following regret.
\begin{definition}\label{def:RBs}
	The oracle that knows the first and last time instances of the $s^{th}$ segment ($T_{s-1}+1$ and $T_s$) and runs the base algorithm between these times will incur the regret $R_{BS.0}(T,S)$ against the competition $\{\vt_{s},t_s\}_{s=1}^S$, which is
	\begin{align*}
		R_{BS.0}(T,S)\leq SR_B\left(\frac{T}{S}\right).
	\end{align*}
	\begin{proof}
		The result comes from the concavity of $R_B(\cdot)$ since
		\begin{align*}
			R_{BS.0}(T,S)\triangleq \sum_{s=1}^{T}R_B(t_s).
		\end{align*}
		where $t_s=T_s-T_{s-1}$ is the length of the $s^{th}$ segment.
	\end{proof}
\end{definition}
Although, we, in general, do not have access to $T_s$, we can utilize the mixability of the loss functions (hence, our mixture framework) to achieve this optimal regret albeit with some redundancy. We begin our design with the expert scheme $\Hau$.

\subsection{Hyper-Expert Scheme}
A straightforward implementation of the mixture framework in Section \ref{sec:mix} is given by creating a number of hyper-experts which run the base algorithm and resets it at every possible switch time. For a $T$ length run of the algorithm, each hyper-expert $i\in\{1,\ldots,N\}$ is defined by a binary sequence of length $T$, where the $t^{th}$ binary value is $1$ if there is a segment starting at $t$ and $0$ otherwise (when there is a segment starting, the base algorithm resets). Thus, we have
\begin{align}
	\Hau_{\text{exp}}: H(i)=\bs{b}_i\triangleq\left\{b_1^{(i)},b_2^{(i)},\ldots,b_t^{(i)},\ldots,b_T^{(i)}\right\}, &&\forall i\label{Hauexp}
\end{align}
where the hyper-expert parameters $b_t^{(i)}$ are such that
\begin{align*}
	b_t^{(i)}\in\{0,1\},&& \forall i,t.
\end{align*} 
Consequently, $b_1$ is always $1$, the number of experts $N$ is $N=2^{T-1}$, and the number of time segments of $i^{th}$ expert is
\begin{align}
	S_i\triangleq\sum_{t=1}^{T}b_t^{(i)}.
\end{align}
\begin{theorem}
	Against a competition of $\{\vt_s^*,t_s\}_{s=1}^S$, we have
	\begin{align}
		E_{S,T}(\Hau_{\text{exp}})\leq& SR_B\left(\frac{T}{S}\right)
	\end{align}
	during time horizon $T$, which is the regret achievable with the base algorithm in Definition \ref{def:RBs}.
	\begin{proof}
	We see that there exists one hyper-expert that resets the base algorithm in the $S$ distinct mutually exclusive time segments (where the optimal parameter remains the same) that divide the time horizon $T$, which provides the result. 
	\end{proof}
\end{theorem}

We observe that our expert creation scheme $\Hau_{\text{exp}}$ has high enough modeling power to achieve the optimal expert regret in Definition \ref{def:RBs}. Next, we design different weighting schemes $\Tau$
and study their redundancies $W_{S,T}(\Tau)$.

\subsection{Weighting Scheme}
We start with a basic design, which is a simple aggregation of all the hyper-experts of the scheme $\Hau_{\text{exp}}$.
\subsubsection{Naive Design}
A direct aggregation of these hyper-experts with uniform priors corresponds to the following:
\begin{align}
\Tau_{\text{exp.n}}:\enspace\tau_t(i,j)=
\displaystyle\begin{dcases}
\begin{aligned}
2^{-T},& &&\text{if } t=1\\
1,& &&\text{if } t\geq 2,i=j\\
0,& &&\text{if } t\geq 2,i\neq j
\end{aligned}
\end{dcases}.\label{transNaive}
\end{align}
\begin{remark}
	$\Tau_{exp.n}$ satisfies Definition \ref{def:Tau} (i.e., bounded by a probability distribution) and is a valid weighting scheme, since there are $2^{T-1}$ experts with initial weights $\tau_1(i,j)=2^{-T}$.
\end{remark}
\begin{proposition}\label{pro:exp.n}
	$\Tau_{\text{exp.n}}$ has the following mixture regret
	\begin{align}
	W_{S,T}(\Tau_{\text{exp.n}})\lessapprox T\log(2)
	\end{align}
	\begin{proof}
		The result comes from the logarithm of $\tau_1(i,j)$.
	\end{proof}
\end{proposition}
Since the weighting is too trivial, the result of Proposition \ref{pro:exp.n} is not sublinear. Next, we provide a more meaningful design.
\subsubsection{Better Design}
A smarter aggregation of these hyper-experts with respect to the number of segments is given by
\begin{align}
\Tau_{\text{exp.b}}:\enspace	\tau_t(i,j)=
\begin{dcases}
\begin{aligned}
\frac{1}{T}\left[\binom{T}{S_j}\right]^{-1},& &&\text{if } t=1\\
1,& &&\text{if } t\geq 2,i=j\\
0,& &&\text{if } t\geq 2,i\neq j
\end{aligned}
\end{dcases}.
\end{align}
\begin{remark}
	$\Tau_{exp.b}$ satisfies Definition \ref{def:Tau} and is a valid weighting scheme, since $\tau_1(i,j)$ is uniform conditioned on $S_j$ and weight of $S_j$ is uniform with $1/T$.
\end{remark}
\begin{proposition}\label{pro:exp.b}
	$\Tau_{\text{exp.b}}$ has the following mixture regret
	\begin{align}
	W_{S,T}(\Tau_{\text{exp.b}})\lessapprox \log(T)+S\log\left(\frac{T}{S}\right)
	\end{align}
	\begin{proof}
		The result comes from $\log(\tau_1(i,j))$ and Lemma \ref{lem:binom}.
	\end{proof}
\end{proposition}
We have a better redundancy in Proposition \ref{pro:exp.b}, where $S\log(T/S)$ and $\log(T)$ result from not knowing the segment start times and the number of segments respectively. However, we can further improve it as follows.
\subsubsection{Optimal Design}
The following setting of the transition weights provides us with an optimal mixture regret.
\begin{align}
\Tau_{\text{exp.o}}:\enspace\tau_t(i,j)=
\begin{dcases}
\begin{aligned}
\left(\frac{2eT}{S_j}\right)^{-S_j},& &&\text{if } t=1\\
1,& &&\text{if } t\geq 2,i=j\\
0,& &&\text{if } t\geq 2,i\neq j
\end{aligned}
\end{dcases}.
\end{align}
\begin{remark}
	$\Tau_{exp.o}$ satisfies Definition \ref{def:Tau} and is a valid weighting scheme from Lemma \ref{lem:binom} and $\sum_{S=1}^{T}2^{-S}< 1, \forall T$.
\end{remark}
\begin{theorem}\label{thm:exp.o}
	$\Tau_{\text{exp.o}}$ has the following mixture regret
	\begin{align}
	W_{S,T}(\Tau_{\text{exp.o}})\lessapprox S\log\left(\frac{T}{S}\right)
	\end{align}
	\begin{proof}
		The result comes from $\log(\tau_1(i,j))$ and $S=o(T)$.
	\end{proof}
\end{theorem}

The mixture regret $W_{S,T}(\Tau_{\text{exp.o}})$ in Theorem \ref{thm:exp.o} corresponds with the minimal redundancy in the source coding \cite{coding1} and will become our benchmark (the optimal mixture redundancy), which we will compare against in the subsequent sections.

\section{A~Tractable~Exact~Approach: A~Polynomial~Time~Complexity~Mixture~Scheme}\label{sec:quad}

While the algorithm in Section \ref{sec:exp} is able to achieve the optimal expert regret, it is unfortunately not a tractable algorithm since the computational complexity is non-polynomial and exponentially grows with time. In this section, we aim to create a tractable alternative.
 
In the design of Section \ref{sec:exp}, we observe that although there are exponential number of possible hyper-experts, they indeed have some overlapping behaviors. Even though, the segment combinations may grow exponentially (since at each time $t$ we may or may not have a new segment starting), the number of possible time segments actually grow quadratically in time (since each time segment $t_s$ starts and finish at specific time instances and the possible combinations of start and finish time are $O(T^2)$). 

Thus, we can create a quadratic time complexity algorithm instead of the exponential time complexity of Section \ref{sec:exp} with the same modeling power. We again begin our design with the hyper-expert scheme.

\subsection{Hyper-Expert Scheme}\label{sec:quad.H}
We construct this algorithm by creating a number of hyper-experts which run the base algorithm in given time intervals such that
\begin{align}
\Hau_{\text{quad}}: H(i)=\{s_i,f_i,l_i\}, &&\forall i\label{Hauquad},
\end{align}
where the hyper-expert parameters
\begin{align*}
	s_{i}\in\{1,2,3,\ldots,T\},
\end{align*}
is the start time (when we start the run of the base algorithm) of the $i^{th}$ expert, 
\begin{align*}
	f_i\in\{s_{i}+1,s_{i}+2,\ldots,T+1\},
\end{align*}
is the finish time (when the base algorithm stops) of the $i^{th}$ expert, and
\begin{align}
	l_i\triangleq f_i-s_i,
\end{align}
is the runtime of the expert $i$. At any given time $t$, we observe that we have at most $T^2/4$ parallel running experts. 
\begin{theorem}
	Against a competition of $\{\vt_s^*,t_s\}_{s=1}^S$, we have
	\begin{align}
	E_{S,T}(\Hau_{\text{quad}})\leq SR_B\left(\frac{T}{S}\right),
	\end{align}
	during time horizon $T$, which is the regret achievable with the base algorithm according to Definition \ref{def:RBs}.
	\begin{proof}
		We see that there exists mixture path of hyper-experts that runs the base algorithm in the $S$ distinct mutually exclusive time segments (where the optimal parameter remains the same) that divide the time horizon $T$. 
	\end{proof}
\end{theorem}

We again observe that our expert creation scheme $\Hau_{\text{quad}}$ is powerful enough to model and achieve the optimal regret in Definition \ref{def:RBs}. Next, we design different weighting schemes $\Tau$
and study their redundancies $W_{S,T}(\Tau)$.

\subsection{Weighting Scheme}
To mix these hyper-experts, we need a weighting scheme as in Definition \ref{def:Tau}. We start by a naive design of the weights.
\subsubsection{Naive Design}
We set the transition weights as 
\begin{align}
\Tau_{\text{quad.n}}:\enspace\tau_t(i,j)=
\begin{dcases}
\begin{aligned}
\frac{1}{T},& &&\text{if } t=f_i=s_j\\
1,& &&\text{if } t<f_i, i=j\\
0,& &&\text{otherwise}
\end{aligned}
\end{dcases},\label{quad.n}
\end{align}
where $f_i=1$ at $t=1$.
\begin{remark}
	$\Tau_{quad.n}$ satisfies Definition \ref{def:Tau} and is a valid weighting scheme, since the total number of hyper-experts that start at time $t$ is bounded by $T$.
\end{remark}
\begin{proposition}\label{pro:quad.n}
	$\Tau_{\text{quad.n}}$ has the following mixture regret
	\begin{align}
		W_{S,T}(\Tau_{\text{quad.n}})\lessapprox S\log(T),
	\end{align}
	\begin{proof}
		From \eqref{quad.n}, we incur $\log(T)$ mixture regret at the start of each segment, which totals up to $S\log(T)$.
	\end{proof}
\end{proposition}
The result in Proposition \ref{pro:quad.n} is a non-optimal mixture regret. Next, we provide a better design to improve the redundancy.
\subsubsection{Better Design}
A smarter design of the weights is
\begin{align}
\Tau_{\text{quad.b}}:\enspace\tau_t(i,j)=
\begin{dcases}
\frac{1}{l_j}-\frac{1}{l_j+1}, &\text{if } t=f_i=s_j\\
1, &\text{if } t<f_i, i=j\\
0, &\text{otherwise}\\
\end{dcases},\label{quad.b}
\end{align}
where $f_i=1$ at $t=1$.
\begin{remark}\label{rem:quad.b}
	$\Tau_{quad.b}$ satisfies Definition \ref{def:Tau} and is a valid weighting scheme; since for every $j$ that has $s_j=f_i$, $l_j$ is distinct and at least $1$. Thus, we have $\sum_{l_j=1}^{T}(l_j)^{-1}-(l_j+1)^{-1}< 1$.
\end{remark}

\begin{proposition}\label{pro:quad.b}
	$\Tau_{\text{quad.b}}$ has the following mixture regret
	\begin{align*}
	W_{S,T}(\Tau_{\text{quad.b}})\lessapprox 2S\log\left(\frac{T}{S}\right),
	\end{align*}
	\begin{proof}
		The proof is in Appendix \ref{app:pro:quad.b}.
	\end{proof}
\end{proposition}
This weighting, which is similar to \cite{willems} and its variants, provides a near-optimal redundancy with a constant multiplicative gap. Next, we provide a weighting with optimal redundancy.
\subsubsection{Optimal Design}We design the transition weights as
\begin{align}
\Tau_{\text{quad.o}}:\enspace\tau_t(i,j)=
\begin{dcases}
\begin{aligned}
\frac{(2l_j)^{-1}}{(\log(el_j))^2},& &&\text{if } t=f_i=s_j\\
1,& &&\text{if } t\neq f_i, i=j\\
0,& &&\text{otherwise}
\end{aligned}
\end{dcases}.\label{quad.o}
\end{align}
where $f_i=1$ at $t=1$.
\begin{remark}\label{rem:quad.o}
	$\Tau_{\text{quad.o}}$ satisfies Definition \ref{def:Tau} and is valid.
	\begin{proof}
		The proof is in Appendix \ref{app:rem:quad.o}.
	\end{proof}
\end{remark}

\begin{theorem}\label{thm:quad.o}
	$\Tau_{\text{quad.o}}$ has the following mixture regret
	\begin{align}
	W_{S,T}(\Tau_{\text{quad.o}})\lessapprox S\log\left(\frac{T}{S}\right)
	\end{align}
	\begin{proof}
		The proof is in Appendix \ref{app:thm:quad.o}.
	\end{proof}
\end{theorem}

The mixture regret of $W_{S,T}(\Tau_{\text{quad.o}})$ in Theorem \ref{thm:quad.o} is the optimal mixture redundancy in Theorem \ref{thm:exp.o}.

\section{A~Practical~Approximate~Approach: A~Logarithmic~Time~Complexity~Algorithm}\label{sec:log}
When we want to compete against the base algorithms that are run separately on the individual time segments of length $t_s$ with optimal competitions $\vt_s^*$; the quadratic time complexity is the best we can do in general since this mixture guarantees that there will at least be one run of the base algorithm in the hyper expert pool that runs between $T_{s-1}+1$ and $T_s$. In some more specific problem settings the knowledge of the stop time of the algorithm may not be needed. However, for a more comprehensive analysis and more general results, we will go with the assumption that we may indeed need the start and stop times of the base algorithm to achieve satisfactory regret results. 

In a situation when the time horizon is unknown the traditional approach is to use the doubling trick \cite{doubling_trick} where we run the algorithm with time lengths that is the double of the previous run (e.g., $2^0, 2^1, 2^2, \ldots$). Incorporating this idea, if we use only the start times in our hyper-expert scheme in Section \ref{sec:quad.H}, we can improve the complexity to linear in time. 

In such a scenario, competing directly against the separate runs of the base algorithm in the individual time segments will not be fair. For a fairer competition, the oracle we want to compete against will not know the start and stop times ($T_{s-1}+1$ and $T_s$ respectively) beforehand, or equivalently, it will not know the stop time $T_s$ of the $s^{th}$ segment at $t=T_{s-1}$. Instead, we compete against an oracle that will know of the stop times $T_{s-1}$ as it comes across that time instance (i.e., the oracle will learn that it needs to start a new segment at $t=T_s+1$).  

\begin{definition}\label{def:RBsD}
	The oracle $BS.1$ that knows the last time of the $s^{th}$ segment $T_{s}$ at $t=T_s$ and runs the base algorithm using the doubling trick will incur the regret
	\begin{align*}
	R_{BS.1}(T,S)\lessapprox&S\log_2\left(\frac{T}{S}\right)R_B\left(\frac{2T}{S\log_2\left(\frac{T}{S}\right)}\right),
	\end{align*}
	from the concavity of $R_B(\cdot)$, where $t_s=T_s-T_{s-1}$ is the length of the $s^{th}$ segment with the competition $\vt_{s}^*$. While tighter regret bounds can be derived for different $R_B(\cdot)$, this bound is the most comprehensive (since it only uses the concavity of the base algorithm regret) and is tight when $R_B(\cdot)= O(1)$.
	\begin{proof}
		The proof is in Appendix \ref{app:def:RBsD}.
	\end{proof}
\end{definition}

The competition and result in Definition \ref{def:RBsD} is intuitive since the straightforward implementation of the doubling technique to the oracle in Section \ref{sec:quad} will create a logarithmic multiplicative redundancy.

We observe that since the hyper-experts use the doubling trick to rerun the base algorithm, they end up running it for lengths that are powers of $2$. Then, we can actually implement our algorithm in a more efficient manner by creating a number of hyper experts that rerun the base algorithm in specific time intervals. These hyper-experts will be designed in such a way that the $i^{th}$ expert, $i\in\{1,\ldots,N\}$, will run the base algorithm individually in subsequent time segments of length $2^k$ for some $k$. 

\subsection{Hyper-Expert Scheme}
We design the hyper-experts as the following
\begin{align}
\Hau_{\text{log}}: H(i)=\{k_i\}, &&\forall i\label{Hauplog},
\end{align}
where $k_i$ is a parameter such that
\begin{align}
	k_i\in\{2^0,2^1,2^2,2^3,\ldots\},
\end{align}
where the hyper-expert $i$ with the parameter $k_i$ start its run at $t=k_i$; and runs the base algorithm for an interval of $k_i$.

\begin{theorem}\label{thm:Hlog}
	Against a competition of $\{\vt_s^*,t_s\}_{s=1}^S$, we have
	at most $S\log_2\left(\frac{8T}{S}\right)$ switches between experts and
	\begin{align}
	E_{S,T}(\Hau_{\text{log}})	\lessapprox&S\log_2\left(\frac{T}{S}\right)R_B\left(\frac{2T}{S\log_2\left(\frac{T}{S}\right)}\right),
	\end{align}
	during time horizon $T$, which is the regret achievable with the base algorithm according to Definition \ref{def:RBsD}.
	\begin{proof}
		The proof is in Appendix \ref{app:thm:Hlog}.
	\end{proof}
\end{theorem}

We observe that our expert creation scheme $\Hau_{\text{log}}$ (which is similar in spirit to \cite{daniely,zhang2019,zhang2020}) is powerful enough to model and achieve the optimal regret in Definition \ref{def:RBsD}. Next, we design different weighting schemes $\Tau$ and study their redundancies $W_{S,T}(\Tau)$.

\subsection{Weighting Scheme}
We start by a standard design of the weights, where we incorporate a time decreasing ($t^{-1}$) switching probability.
\subsubsection{Naive Design}
One of the most traditional ways of setting the switching probability is the following:
\begin{align}
\Tau_{\text{log.n}}:\enspace\tau_t(i,j)=
\begin{dcases}
\begin{aligned}
1,& &&\text{if } t=1,k_i=2^0\\
\frac{t-1}{t},& &&\text{if } t\geq 2, i=j\\
\frac{1}{t\log_2(2t)},& &&\text{if } t\geq 2,t\geq k_j, i\neq j\\
0,& &&\text{otherwise}
\end{aligned}
\end{dcases},\label{log.n}
\end{align}
\begin{remark}\label{rem:log.n}
	$\Tau_{log.n}$ satisfies Definition \ref{def:Tau} and is valid since there is at most $\log_2(2t)$ experts running at $t$.
\end{remark}
\begin{proposition}\label{pro:log.n}
	$\Tau_{\text{log.n}}$ has the following mixture regret
	\begin{align*}
	W_{S,T}(\Tau_{\text{log.n}})\lessapprox 2S\log_2\left(\frac{T}{S}\right)\log(T),
	\end{align*}
	\begin{proof}
		If there is a switch at $t=t_i$, we will, at most, incur 
		\begin{align*}
			2\log(t_i)+\log\log_2(2t_i)\leq&2\log(T)+\log\log_2(2T),
		\end{align*}
		for a time horizon $T$. Since, we have at most $S\log_2\left(\frac{8T}{S}\right)$ switches, the redundancy becomes
		\begin{align*}
			W_{S,T}(\Tau_{\text{log.n}})\lessapprox 2S\log_2\left(\frac{T}{S}\right)\log(T),
		\end{align*}
		which concludes the proof.
	\end{proof}
\end{proposition}
The result in Proposition \ref{pro:log.n} is non-optimal with a logarithmic optimality gap. Next, we make the switching probability dependent on the experts individual runtime instead of the global runtime $t$, to improve the optimality gap.
\subsubsection{Better Design}
A better design is as follows:
\begin{align}
\Tau_{\text{log.b}}:\enspace\tau_t(i,j)=
\begin{dcases}
\begin{aligned}
1,& &&\text{if } t=1,k_i=2^0\\
\frac{z_{i,t}-1}{z_{i,t}},& &&\text{if } t\geq 2,z_{i,t}\neq1, i=j\\
\frac{1}{\log_2(2t)},& &&\text{if } t\geq 2,z_{i,t}=1, i=j\\
\frac{1}{z_{i,t}\log_2(2t)},& &&\text{if } t\geq 2, t\geq k_j, i\neq j\\
0,& &&\text{otherwise}
\end{aligned}
\end{dcases},\label{log.b}
\end{align}
where $z_{i,t}$ is the current round of the $i^{th}$ expert at time $t$. As an example, the expert with $k_i=2^2$, will have $\{z_{i,t}\}_{t=1}^T=\{-,-,-,1,2,3,4,1,2,3,4,1,2,3,4,\ldots\}$. 
\begin{remark}\label{rem:log.b}
	$\Tau_{log.b}$ satisfies Definition \ref{def:Tau} and is valid since there is at most $\log_2(2t)$ experts running at $t$. 
\end{remark}
\begin{proposition}\label{pro:log.b}
	$\Tau_{\text{log.b}}$ has the following mixture regret
	\begin{align*}
	W_{S,T}(\Tau_{\text{log.b}})\lessapprox S\log_2\left(\frac{T}{S}\right)\log\left(\frac{T{\log_2(T)}}{S\log_2(T/S)}\right)
	\end{align*}
	\begin{proof}
		For each segment of length $t_s$, we will incur 
		\begin{align*}
		\log(t_s)+\log\log_2(2t)\leq&\log(t_s)+\log\log_2(2T),
		\end{align*}
		for a time horizon $T$. Since, we have at most $S\log_2\left(\frac{8T}{S}\right)$ segments, we get the result.
	\end{proof}
\end{proposition}
While the result in Proposition \ref{pro:log.b} is an improvement, we still have a logarithmic (albeit better) optimality gap. Next, we utilize the structure of the hyper-expert scheme and limit the times the experts can switch to improve the optimality gap.
\subsubsection{Smarter Design}\label{sec:log.s}
When we utilize the hyper-expert scheme, we get the following weighting.
\begin{align}
\Tau_{\text{log.s}}:\enspace\tau_t(i,j)=
\begin{dcases}
\begin{aligned}
1,& &&\text{if } t=1,k_i=2^0\\
1,& &&\text{if } t\geq 2, z_{i,t+1}\neq 1, i=j\\
\frac{1}{\log_2(2t)},& &&\text{if } t\geq 2, z_{i,t+1}=1\\
0,& &&\text{otherwise}
\end{aligned}
\end{dcases},\label{log.s}
\end{align}
where $z_{i,t}$ is the current round of the $i^{th}$ expert at time $t$.
\begin{remark}\label{rem:log.s}
	$\Tau_{log.s}$ satisfies Definition \ref{def:Tau} and is valid since there is at most $\log_2(2t)$ experts running at $t$. 
\end{remark}
\begin{proposition}\label{pro:log.s}
	$\Tau_{\text{log.s}}$ has the following mixture regret
	\begin{align*}
	W_{S,T}(\Tau_{\text{log.s}})\lessapprox S\log_2\left(\frac{T}{S}\right)\log\log_2T
	\end{align*}
	\begin{proof}
		We incur $\log\log_2(2T)$ redundancy for every segment, which is at most $S\log_2\left(\frac{8T}{S}\right)$, and concludes the proof.
	\end{proof}
\end{proposition}
With this design, we further improved the optimality gap to the doubly logarithmic factors albeit still non-optimal. Finally, we propose the following weighting, which improves the optimality gap up to constant factors, thus achieves a near-optimal redundancy.
\subsubsection{Optimal Design}
We design the transition weights as the following to achieve a constant optimality gap instead of a divergent one.
\begin{align}
\Tau_{\text{log.o}}:\enspace\tau_t(i,j)=
\begin{dcases}
\begin{aligned}
1,& &&\text{if } t=1,k_i=2^0\\
1,& &&\text{if } t\geq 2, z_{i,t+1}\neq 1, i=j\\
\frac{k_j}{2g_t},& &&\text{if } t\geq 2, z_{i,t+1}=1\\
0,& &&\text{otherwise}
\end{aligned}
\end{dcases},\label{log.o}
\end{align}
where $z_{i,t}$ is the current round of the $i^{th}$ expert at time $t$; and $g_t=\max_{i:z_{i,t=1}} k_i$.
\begin{remark}\label{rem:log.o}
	$\Tau_{log.o}$ satisfies Definition \ref{def:Tau} and is valid since if the $i^{th}$ expert with $k_{i}$ has $z_{i,t}=1$, all $k_j=2^{-n} k_i$ for some $n\geq0$ also has $z_{j,t}=1$.
\end{remark}
\begin{theorem}\label{thm:log.o}
	$\Tau_{\text{log.o}}$ has the following mixture regret
	\begin{align*}
	W_{S,T}(\Tau_{\text{log.o}})\lessapprox 2S\log\left(\frac{T}{S}\right),
	\end{align*}
	\begin{proof}
		The proof is in Appendix \ref{app:thm:log.o}.
	\end{proof}
\end{theorem}
The result in Theorem \ref{thm:log.o} is near optimal. It has an optimality gap of $2$ when compared against the optimal redundancy of the brute force approach in Theorem \ref{thm:exp.o}. By making the transition probabilities dependent on the length of the runtime of the base algorithm as we did in Section \ref{sec:quad} whilst also limiting the transitions to certain times, we have eliminated the divergent optimality gap (specifically the doubly logarithmic gap in Proposition \ref{pro:log.s}). Hence, we have successfully achieved a near optimal mixture using an efficient algorithm with only logarithmic per time complexity.

\section{Conclusion}\label{sec:con}
We investigated the problem of online learning under mixable losses. While algorithms that can compete against the best fixed estimation in hindsight are abundant in the literature. We have investigated this problem in the more general setting of competing against a best switching estimation in hindsight.
To solve the problem, we have introduced techniques (meta algorithms), which can utilize the algorithms that can compete against the best fixed estimation (which we call the base algorithm). Our algorithms consist of two stages of design, which are the hyper-expert schemes (which decide how the base algorithms are run) and the weighting schemes (which decide how the estimations are combined or aggregated together). Our algorithms asymptotically achieve the performance of the best dynamic estimation sequence with near optimal regret redundancies. We proposed two mixtures in our work. Firstly, we proposed a tractable polynomial time complexity algorithm that can achieve optimal mixture regret (or redundancy). Secondly, we proposed an efficient logarithmic time complexity algorithm that can achieve the optimal redundancy up to a constant multiplicity gap. Our regret bounds are guaranteed to hold in a strong deterministic sense in an individual sequence manner.

\bibliographystyle{IEEEtran}
\bibliography{double_bib}

\begin{appendices}
	\section{Proof of Theorem \ref{thm:mixTrans}}\label{app:lem:mixTrans}
	Using \eqref{thetahat} and the mixability in Definition \ref{def:mixable}, we can write
	\begin{align}
	e^{-\alpha l_t(\hat{\bs{\theta}}_t)} \geq \sum_{i\in\Nau_t}^{} P_{i,t} e^{-\alpha l_t(\bs{\theta}_{i,t})}.
	\end{align}
	By taking the logarithm of both sides and dividing by $(-\alpha)$, we acquire the following upper bound on the loss we incur at time $t$
	\begin{align}
	l_t(\hat{\bs{\theta}}_t) \leq - \frac{1}{\alpha} \log{\left( \sum_{i\in\Nau_t}^{} P_{i,t} e^{-\alpha l_t(\bs{\theta}_{i,t})} \right)}.\label{lt1}
	\end{align}
	We use the definition in \eqref{Pit} inside the logarithm in \eqref{lt1} to get
	\begin{align}
	\sum_{i\in\Nau_t}^{} P_{i,t} e^{-\alpha l_t(\bs{\theta}_{i,t})}=\displaystyle\frac{\sum_{i\in\Nau_t}^{} \widetilde{P}_{i,t} e^{-\alpha l_t(\bs{\theta}_{i,t})}}{\sum_{i\in\Nau_t}^{} \widetilde{P}_{i,t}}.\label{laa1}
	\end{align}
	We use \eqref{PitRec} and Definition \ref{def:Tau} in \eqref{laa1} to get
	\begin{align}
	\sum_{i\in\Nau_t}^{}& P_{i,t} e^{-\alpha l_t(\bs{\theta}_{i,t})}\nonumber
	\\&= \frac{\sum_{i\in\Nau_t}^{} \widetilde{P}_{i,t} e^{-\alpha l_t(\bs{\theta}_{i,t})}}{\sum_{i\in\Nau_t}^{} \sum_{j\in\Nau_{t-1}}^{} \widetilde{P}_{j,t-1} e^{-\alpha l_{t-1}(\bs{\theta}_{j,t-1})} \tau_t(j,i)},\nonumber\\
	&\geq\frac{\sum_{i\in\Nau_t}^{} \widetilde{P}_{i,t} e^{-\alpha l_t(\bs{\theta}_{i,t})}}{\sum_{j\in\Nau_{t-1}}^{} \widetilde{P}_{j,t-1} e^{-\alpha l_{t-1}(\bs{\theta}_{j,t-1})}}\label{PeSum1}.
	\end{align}
	Observe that the denominator of \eqref{PeSum1} at time $t$ corresponds to the numerator of \eqref{PeSum1} at time $t-1$.
	Hence, the sum of \eqref{lt1} from $t=1$ to $T$ is upper bounded by 
	\begin{align}
	\sum_{t=1}^{T} l_t(\hat{\bs{\theta}}_t) \leq -\frac{1}{\alpha} \log\left(\sum_{i\in\Nau_T}^{} \widetilde{P}_{i,T}e^{-\alpha l_T(\bs{\theta}_{i,T})} \right)\label{ltSum1}.
	\end{align}
	We point out that because of the recursive calculation of $\widetilde{P}_{i,t}$ coming from all possible algorithm transitions, the sum inside the logarithm in \eqref{ltSum1} includes all possible algorithm transitions. Since $-\log(\cdot)$ is a decreasing function, the total incurred loss is upper bounded by
	\begin{align}
	\sum_{t=1}^T l_t(\hat{\bs{\theta}}_t) \leq -\frac{1}{\alpha}\log \left( \prod_{t=1}^T e^{-\alpha l_t(\bs{\theta}_{I_t,t})} \left(\prod_{t=1}^T \tau_t(I_{t-1},I_t) \right)\right).
	\end{align}
	for some index set $\{I_1,I_2,\ldots,I_T\}$, where $\tau_1(I_0,I_1)\leq P_{I_1,1}$.
	Thus, we have
	\begin{align} \label{lossbound1}
	\sum_{t=1}^T l_t(\hat{\bs{\theta}}_t) \leq \sum_{t=1}^T l_t(\bs{\theta}_{I_t,t})-\frac{1}{\alpha}\log \left(\prod_{t=1}^T \tau_t(I_{t-1},I_t) \right),
	\end{align}
	which concludes the proof.
	
	\section{Proof of Proposition \ref{pro:quad.b}}\label{app:pro:quad.b}
	Let $t_s$ be the length of the time segments where the competition stays the same. Then, from \eqref{quad.b}, we have
	\begin{align}
	W_{S,T}(\Tau_{\text{quad.b}})=&\max_{\{t_s\}_{s=1}^S}\sum_{s=1}^S\left(\log(t_s)+\log(t_s+1)\right)\\
	\leq&S\log\left(\frac{T}{S}\right)+S\log\left(\frac{T}{S}+1\right)\\
	\lessapprox& 2S\log\left(\frac{T}{S}\right),
	\end{align}
	which concludes the proof.
	
	\section{Proof of Remark \ref{rem:quad.o}}\label{app:rem:quad.o}
	For every $j$ that has $s_j=f_i$, $l_j$ is distinct and at least $1$. Then, we have
	\begin{align}
	\sum_{j}\tau_{f_i}(i,j)\leq&\sum_{l_j=1}^{T}\frac{(2l_j)^{-1}}{(\log(l_j)+1)^2}\\
	\leq&\left.\frac{(2x)^{-1}}{(\log(x)+1)^2}\right\rvert_{x=1}+\int_{1}^T\frac{(2x)^{-1}}{(\log(x)+1)^2}\mathrm{d}x,
	\end{align}
	from the fact that ${(2l_j)^{-1}}{(\log(l_j)+1)^{-2}}$ is decreasing for $l_j\geq 1$. Hence, we get
	\begin{align}	
	\sum_{j}\tau_t(i,j)\leq&\frac{1}{2}-\left.\frac{1}{2(\log(x)+1)}\right\rvert_1^T\\
	\leq&1,
	\end{align}
	which concludes the proof.
	
	\section{Proof of Theorem \ref{thm:quad.o}}\label{app:thm:quad.o}
	Let $t_s$ be the length of the time segments where the competition stays the same. Then, from \eqref{quad.o}, we have
	\begin{align}
	W_{S,T}(\Tau_{\text{quad.o}})=&\max_{\{t_s\}_{s=1}^S}\sum_{s=1}^S\log(2t_s)+2\log(\log(t_s)+1)\\
	\leq&S\log\left(\frac{2T}{S}\right)+2S\log\left(\log\left(\frac{T}{S}\right)+1\right),
	\end{align}
	from the concavity. Hence, we get
	\begin{align}
	W_{S,T}(\Tau_{\text{quad.o}})\lessapprox& S\log\left(\frac{T}{S}\right),
	\end{align}
	since $S\leq o(T)$, which concludes the proof.
	
	\section{Proof of Definition \ref{def:RBsD}}\label{app:def:RBsD}
	We have $$\sum_{n=0}^{N_s-1}2^{n}+t_s'=t_s\leq\sum_{n=0}^{N_s}2^{n},$$ where the oracle resets $N_s$ times in the $s^{th}$ segment. Thus
	\begin{align}
	2^{N_s}\leq t_s\leq 2^{N_s+1}-1\leq2t_s-1.\label{t_s}
	\end{align}
	Using \eqref{t_s}, we have the following
	\begin{align*}
	R_{BS.1}(T,S)\leq&\sum_{s=1}^{S}\left(\sum_{n=0}^{N_s}R_B(2^n)\right)\\
	\leq& \left(\sum_{s=1}^{S}(N_s+1)\right)R_B\left(\frac{\sum_{s=1}^{S}(2t_s-1)}{\sum_{s=1}^{S}(N_s+1)}\right)\\
	\leq&\left(\sum_{s=1}^{S}\log_2(2t_s)\right)R_B\left(\frac{2T-S}{\sum_{s=1}^{S}\log_2(t_s+1)}\right)\\
	\leq&SR_B\left(\frac{2T-S}{S-1+\log_2(T-S+2)}\right)\\
	&+\left({\sum_{s=1}^{S}\log_2(t_s)}\right)R_B\left(\frac{2T-S}{{\sum_{s=1}^{S}\log_2(t_s)}}\right).\\
	\end{align*}
	Using $S\leq o(T)$ and the concavity of $R_B(\cdot)$, we have
	\begin{align*}
	R_{BS.1}(T,S)\leq&SR_B\left(\frac{2T}{S}\right)+\hat{S}R_B\left(\frac{2T}{\hat{S}}\right)\\
	\lessapprox&\hat{S}R_B\left(\frac{2T}{\hat{S}}\right),
	\end{align*}
	where $\hat{S}=S\log_2\left(\frac{T}{S}\right)$, and concludes the proof.
	
	\section{Proof of Theorem \ref{thm:Hlog}}\label{app:thm:Hlog}
	Since the competition is against $S$ segments, there will be $S-1$ changes during the time horizon to compete against a different parameter. At these changes, we need a new base algorithm starting for satisfactory competition. We can consider the start time $t=0$ as a dummy change. We observe that in our hyper-expert pool, if there is a hyper expert that runs the base algorithm in a certain interval with length at least $2$, we also have an hyper expert that runs the base algorithm in the first and the second half of that interval individually. If there exists an optimal parameter change (including the dummy change) in a particular time segment of an hyper-expert, we need to utilize the hyper-expert with the smaller consecutive segments, which are the first and second half of that particular segment, since the goal is to extract time segments without any in-segment parameter change. Thus, we will split the time from $t=1$ to $T$ in such a way that no time segment will include a parameter change. For the regret of this hyper-expert scheme, we need to bound the final number of these time segments. 
	We point out that whenever we utilize the first and second half of a particular segment, we increase the total number of segments our base algorithm runs in by $1$. 
	
	Let the time horizon be $2^{N-1}\leq T\leq2^N-1$, where the largest implicit segment is of length $2^N$ (from $0$ to $2^N-1$). Starting from the first change (the dummy change at $t=0$), we will split the top segment and continue splitting the first half of each segment. This will increase the number of segments by $N$. For the second change, we can increase the number of segments at most by $N-1$ since the largest segment that has not been split is of length $2^{N-1}$. After that, we have two segments of length $2^{N-2}$ that has not been split left. Hence, the number of segments is bounded by
	\begin{align}
	\tilde{S}\leq &N+\sum_{x=1}^{a}2^{x-1}(N-x)+K(N-a-1),
	\end{align} 
	for some $K$ and $a$, where $2^a+K=S$ and $K\leq 2^a$. Thus,
	\begin{align}
	\tilde{S}\leq SN-\sum_{x=1}^{a+1}(S-2^{x-1}).
	\end{align}
	Since $K\leq2^a\leq K+2^a=S\leq2^{a+1}\leq 2S$, we have
	\begin{align}
	\tilde{S}&\leq S(N-(a+1))+2^{a+1}\\
	&\leq S\log_2\left(\frac{2T}{2^{a+1}}\right)+2^{a+1}\\
	&\leq S\log_2\left(\frac{8T}{S}\right)
	\end{align}
	Hence, from the concavity of $R_B(\cdot)$, we have
	\begin{align}
	E_{S,T}(\Hau_{\text{log}})	\leq&S\log_2\left(\frac{8T}{S}\right)R_B\left(\frac{2T}{S\log_2\left(\frac{8T}{S}\right)}\right),\\
	\lessapprox&S\log_2\left(\frac{T}{S}\right)R_B\left(\frac{2T}{S\log_2\left(\frac{T}{S}\right)}\right),
	\end{align}
	which concludes the proof.	
	
	\section{Proof of Theorem \ref{thm:log.o}}\label{app:thm:log.o}
	From \eqref{log.o}, we see that at every switch we incur $(m+1)\log(2)$ redundancy if we switch to the expert with $k_i=2^{-m}g_t$, where $g_t$ is the largest parameter that restarts the base algorithm at $t$. In a sense, we incur $\log(2)$ regret whenever we switch from $k_i$ to $k_i/2$. Moreover, since another segment with $k_i/2$ starts in the second half of the segment with $k_i$, we will additionally incur another $\log(2)$ regret because $k_i/2$ is the largest parameter at the middle of the segment with $k_i$. Thus, every segment costs us $\log(4)$ regret, which gives
	\begin{align}
	W_{S,T}(\Tau_{\text{log.o}})&\leq S\log_2\left(\frac{8T}{S}\right)\log(4)\\
	&\leq 2S\log\left(\frac{8T}{S}\right)\\
	&\lessapprox 2S\log\left(\frac{T}{S}\right),
	\end{align} 
	which concludes the proof.
\end{appendices}

\end{document}